\documentclass{article}

\usepackage[preprint]{neurips_2024} 

\usepackage[utf8]{inputenc} 
\usepackage[T1]{fontenc}    
\usepackage{hyperref}       
\usepackage{url}            
\usepackage{booktabs}       
\usepackage{amsfonts}       
\usepackage{nicefrac}       
\usepackage{microtype}      
\usepackage{xcolor}         

\usepackage{multirow}
\usepackage{graphicx}
\usepackage{makecell}
\usepackage{colortbl}
\usepackage{longtable}
\usepackage{comment}
\usepackage{caption}
\usepackage{tcolorbox}
\usepackage{subcaption}
\usepackage{booktabs} 
\usepackage{wrapfig}
\usepackage{xparse}

\definecolor{lightyellow}{rgb}{1, 0.95, 0.85}
\definecolor{graphicbackground}{rgb}{0.9765,0.9451,0.9059}
\definecolor{codebackground}{rgb}{0.8314,0.949,0.9882}

\title{Building and better understanding vision-language models: insights and future directions}

\author{
\textbf{Hugo Laurençon}\thanks{Equal contribution} \quad \textbf{Andrés Marafioti} \quad \textbf{Victor Sanh} \quad \textbf{Léo Tronchon}$^{*}$\vspace{1em}\\
Hugging Face
}

\begin{document}

\maketitle

\begin{abstract}
The field of vision-language models (VLMs), which take images and texts as inputs and output texts, is rapidly evolving and has yet to reach consensus on several key aspects of the development pipeline, including data, architecture, and training methods. This paper can be seen as a tutorial for building a VLM. We begin by providing a comprehensive overview of the current state-of-the-art approaches, highlighting the strengths and weaknesses of each, addressing the major challenges in the field, and suggesting promising research directions for underexplored areas. We then walk through the practical steps to build Idefics3-8B, a powerful VLM that significantly outperforms its predecessor Idefics2-8B, while being trained efficiently, exclusively on open datasets, and using a straightforward pipeline. These steps include the creation of Docmatix, a dataset for improving document understanding capabilities, which is 240 times larger than previously available datasets. We release the model along with the datasets created for its training.
\end{abstract}

\section{Introduction}

Vision-language models (VLMs), that take images and texts as inputs and output texts, are highly effective in various applications such as document and figure understanding \citep{mPLUG-DocOwl-1.5}, solving visual mathematical problems \citep{G-llava-Geo170K}, or converting webpage screenshots into code \citep{WebSight}. The advancement of powerful open large language models \citep{Llama2, Mistral7B, Gemma} and vision encoders \citep{SigLIP, EVA-CLIP, CLIP} allows researchers to build upon these unimodal pre-trained models to create advanced VLMs that solve these tasks with increasing accuracy \citep{InstructBLIP, LLaVA, Qwen-VL, SPHINX, Monkey, CogVLM}.

Despite advancements in the field, the literature highlights a variety of divergent design choices across key aspects of the development pipeline, indicating a lack of consensus. For instance, while many recent models \citep{FROMAGe,BLIP-2,LLaVA} have chosen to concatenate the sequence of image hidden states with the sequence of text embeddings before feeding it as input to the language model, the Llama 3-V model \citep{Llama3} use interleaved Transformer-based cross-attentions to fuse the visual information into the LLM, similar to Flamingo \citep{Flamingo}. These different core choices in VLM development, often not ablated or justified in research papers, make it challenging to distinguish which decisions impact model performance and assess the compute and data efficiency trade-offs associated with each method.

In this paper, we begin by guiding the reader through the main research questions in the field, offering a detailed overview of the latest VLM approaches to address these challenges, along with the strengths and weaknesses of each. Specifically, we focus on (a) the various architectures used to connect pre-trained language models with vision encoders, (b) the different types of data employed in VLM training, their utility, and the typical stage at which they are introduced, (c) the training methods for VLMs, which are often divided into multiple stages for efficiency and stability, and (d) the challenges encountered in model evaluation. We propose future research directions, particularly around data, to enhance model performance.

Building on this overview, we then walk through the practical steps for building Idefics3-8B\footnote{\url{https://huggingface.co/HuggingFaceM4/Idefics3-8B-Llama3}}, a powerful VLM trained efficiently, using only open datasets and a straightforward pipeline. Idefics3-8B significantly outperforms its predecessor, Idefics2-8B, particularly in document understanding tasks, with a 13.7-point improvement on DocVQA \citep{DocVQA}. To especially boost the capabilities on this task, we created the Docmatix\footnote{\url{https://huggingface.co/datasets/HuggingFaceM4/Docmatix}} dataset, which includes 2.4 million images and 9.5 million QA pairs derived from 1.3 million PDF documents—a 240-fold increase in scale compared to previous open datasets. We release our model alongside the datasets used for its training.

\section{Analyzing architectural choices in VLMs}

\subsection{Connecting unimodal pre-trained models}

Since the introduction of Frozen \citep{frozen} and Flamingo \citep{Flamingo}, most VLMs have been built on top of unimodal pre-trained backbones, a language model and/or a vision encoder, rather than training entirely new models from scratch \citep{FROMAGe, BLIP-2, LLaVA}. The availability of powerful open-source LLMs \citep{Llama3, Mistral7B, Gemma} and image encoders \citep{SigLIP, EVA-CLIP, CLIP}, which are increasingly expensive to train, enables researchers to leverage these models to create high-performing VLMs at a reduced cost \citep{InstructBLIP, FROMAGe, LLaVA, DePALM}. These two pre-trained models are usually connected with either a cross-attention or a self-attention architecture.

\subsubsection{Cross-attention architecture}

The cross-attention architecture is introduced in Flamingo \citep{Flamingo}. The image hidden states encoded by the vision backbone are used to condition the frozen language model using freshly initialized cross-attention layers that are interleaved between the pretrained language model layers. The keys and values in these layers are obtained from the vision features, while the queries are derived from the language inputs. In practice, a cross-attention block is inserted after every four Transformer blocks in the LLM, adding newly initialized parameters equivalent to roughly 1/4th of the LLM's size. This significant increase in parameters enhances the model's expressivity, allowing it to achieve strong performance without unfreezing the LLM during training, thereby preserving the pre-trained LLM's performance on text-only tasks.

Idefics1 \citep{OBELICS} and OpenFlamingo \citep{OpenFlamingo} are open replications of Flamingo. More recently, Llama 3-V \citep{Llama3} also adopted this approach to adapt Llama 3 to multimodality.

\subsubsection{Self-attention architecture}

\begin{figure}[ht]
\centering
\includegraphics[width=0.9\textwidth]{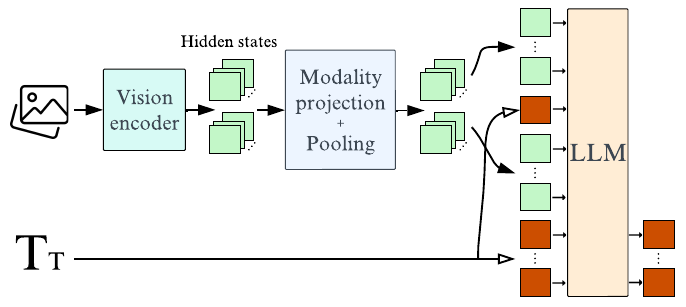}
\caption{From \cite{Idefics2}. The self-attention, or fully-autoregressive, architecture: Input images are processed by the Vision encoder. The resulting visual features are mapped (and optionally pooled) to the $LLM$ input space to get the visual tokens. They are concatenated (and potentially interleaved) with the input sequence of text embeddings (green and red column). The concatenated sequence is fed to the language model ($LLM$), which predicts the text tokens output.}
\label{fig:self_attention_architecture}
\end{figure}

In the self-attention architecture (or fully-autoregressive architecture), introduced in FROMAGe \citep{FROMAGe} and BLIP2 \citep{BLIP-2}, the output of the vision encoder is treated as tokens and concatenated to the sequence of text tokens. The entire sequence is then passed as input to the language model. The sequence of visual tokens can be optionally pooled into a shorter sequence, making the model more efficient both during the training and at inference. We refer to the layers that map the vision-hidden space to the text-hidden space as modality projection layers. Figure \ref{fig:self_attention_architecture} highlights the different components of the self-attention architecture.

Most recent VLMs have now adopted this design, including Llava \citep{LLaVA}, Qwen-VL \citep{Qwen-VL}, DeepSeek-VL \citep{DeepSeek-VL}, SPHINX \citep{SPHINX}, VILA \citep{VILA}, MiniGemini \citep{Mini-Gemini}, Monkey \citep{Monkey}, MM1 \citep{MM1}, Idefics2 \citep{Idefics2}, MiniCPM-V \citep{MiniCPM-V}, InternLM \citep{Internlm-xcomposer2-4khd} or InternVL \citep{InternVL}.

\subsubsection{Which architecture performs best?}

The performance comparison between these two main types of architectures was explored in \cite{Idefics2}. The pre-trained unimodal models are Mistral-7B \citep{Mistral7B} for the LLM and SigLIP-SO400M \citep{SigLIP} for the vision encoder. The model with the self-attention architecture has a total of 8.3B parameters, including 740M newly initialized, while the model with the cross-attention architecture has a total of 10B parameters, including 2.5B newly initialized. The authors demonstrate that the cross-attention architecture significantly outperforms when the backbones are kept frozen during training. However, when parts of the vision encoder and language model are trained with LoRA \citep{LoRA}, adding an extra 200M trainable parameters distributed across both models, the cross-attention architecture performs worse despite having more parameters overall.

Nonetheless, this study did not evaluate the performance of the VLMs on text-only benchmarks. Intuitively, when parts of the language model are unfrozen during training, we need to incorporate data from the LLM training data mixture into the VLM training data to maintain performance on text-only benchmarks.

\subsubsection{Impact of the pre-trained backbones on performance}

Various studies find that the performance of each standalone unimodal pre-trained backbone correlates with the performance of the resulting VLM. For instance, in \citep{Idefics2}, the authors demonstrate that replacing the language model from LLaMA-1-7B \citep{LLaMA} (35.1\% on MMLU \citep{MMLU}) with Mistral-7B \citep{Mistral7B} (60.1\% on MMLU) leads to a substantial improvement across benchmarks. Analogously, replacing CLIP-ViT-H \citep{CLIP} (78.0\% on ImageNet \citep{ImageNet}) with SigLIP-SO400M \citep{SigLIP} (83.2\% on ImageNet), also leads to a substantial performance improvement across all benchmarks, without changing the total number of parameters of the VLM.

Because vision encoders are often trained on different datasets and optimized for various tasks, some models, like SPHINX \citep{SPHINX}, combine representations from multiple encoders, such as DINOv2 \citep{DINOv2} and CLIP \citep{CLIP}, to create a richer sequence of visual embeddings, though this comes at the expense of computational efficiency.

Recent research has heavily focused on improving open language models \citep{LLaMA, Llama3, Gemma, Mistral7B, Vicuna, Dolly, OpenELM, Phi-3, MiniCPM, DeepSeek, Qwen}. In contrast, few open-vision encoders have been released, with SigLIP-SO400M standing out due to its favorable performance-to-parameter ratio with only 400M parameters. This suggests a need for extensively trained open-source vision encoders at scale.

\subsection{Examining the other architectural choices}

\subsubsection{Is a vision encoder really necessary?}

Instead of employing a vision encoder, Fuyu \citep{fuyu} feeds image patches directly into the language model after applying a simple linear projection to adjust the dimensions. This architecture offers two main advantages: it is independent of another pre-trained model and preserves all the information from the original image. The latter point is crucial since the original image details might be necessary for accurately responding to the prompt. On the other hand, a pre-trained vision encoder transforms an image into a representation that is independent of the user's prompt. As a result, vision encoders aim to capture as much information as possible and can still miss details pertinent to the prompt. VisFocus \citep{VisFocus} attempts to address this drawback by incorporating the user's prompt into the vision encoder. However, this approach is less natural in interleaved image-text conversations, where prompts may refer back to previous questions.

Despite these advantages, this architecture has not yet demonstrated superior performance. Fuyu scores significantly lower on benchmarks compared to the best models of similar size released around the same time. PaliGemma \citep{PaliGemma} also experimented with this approach and reported a notable drop in performance compared to using a pre-trained vision encoder. The authors suggest that bypassing a vision encoder pre-trained on billions of images could lead to longer training times to achieve similar performance.\newline 
Furthermore, handling image representation within the language model might decrease its performance on text-only benchmarks. Even if this approach outperformed others on multimodal benchmarks, most VLMs are still not evaluated on text-only benchmarks, making it unclear whether omitting a vision encoder affects text benchmark performance.\newline
Finally, this approach has not been tested yet with an efficient pooling strategy that does not significantly reduce information by operating directly on raw pixels. Looking ahead, for tasks like video understanding or extension to other modalities, it will be important to develop an architecture that can efficiently reduce the number of visual tokens passed to the language model to maintain a reasonable sequence length.

\subsubsection{How should we connect the vision encoder to the language model?}

Many models, such as FROMAGe \citep{FROMAGe} and LLaVA \citep{LLaVA}, use a simple linear layer between the vision encoder and the LLM, ensuring that all encoded visual information is retained since no pooling strategy is applied. However, this approach results in a long sequence of visual tokens, making training and inference less efficient. To address this, Qwen-VL \citep{Qwen-VL} reduces the number of visual tokens by using a single-layer cross-attention module between a group of embeddings and the image hidden states. Similarly, Idefics2 \citep{Idefics2} employs a cross-attention module within a perceiver resampler \citep{perceiver, Flamingo}, demonstrating that the number of visual tokens can be compressed to as few as 64 (divided by 77) while maintaining performance for most tasks, except those that require extensive OCR capabilities. InternLM-XComposer2-4KHD \citep{Internlm-xcomposer2-4khd} also shows that increasing the number of visual tokens per image is primarily necessary for benchmarks focused on OCR tasks, such as InfoVQA \citep{InfographicVQA} and DocVQA \citep{DocVQA}.

Despite the efficiency of the perceiver resampler, its use has been challenged in several papers, which suggest leveraging the 2D structure of images more effectively. For instance, HoneyBee \citep{Honeybee} introduces the C-Abstractor, which reintroduces 2D positional embeddings to the visual features, followed by ResNet blocks \citep{ResNet}. In mPLUG-DocOwl-1.5 \citep{mPLUG-DocOwl-1.5}, the H-Reducer is introduced, using convolutions to divide the number of image hidden states by 4. InternVL \citep{InternVL} also achieves a fourfold compression using a simple pixel shuffle strategy. Recently, MiniCPM-V 2.6 \citep{MiniCPM-V}, like Idefics2, chose the perceiver resampler with 64 learnable embeddings but enhanced it by adding 2D positional embeddings.

\subsubsection{The image-splitting strategy: a trick to increase the number of visual tokens}

Introduced in UReader \citep{UReader} and SPHINX \citep{SPHINX}, the image splitting strategy involves dividing an original image into multiple sub-images, each of which is encoded separately by the vision encoder. The number of tiles can be fixed, such as consistently using four crops per image, or it can vary depending on the image’s original resolution, with the image split every N pixels, for example.

When the number of tiles is based on the original resolution, the model is trained with varying numbers of visual tokens. This approach is particularly advantageous during inference: for simpler tasks, fewer visual tokens are needed, saving computational resources, while more computing can be allocated by increasing the image resolution for tasks that require intensive OCR. This flexibility is highly beneficial for models designed to excel both at reasoning on a single image with high computational resources and at processing videos with many frames while maintaining a reasonable sequence length by using a lower resolution for each frame.

Most vision encoders are designed for relatively low, fixed image resolutions and are not well-suited for processing large images. The image-splitting strategy addresses this by enabling the use of off-the-shelf pre-trained vision encoders at their original resolution, simply by feeding multiple smaller sub-images to the encoder instead of the original large image. Since the vision encoder's weights are shared across each sub-image, this approach also enhances training efficiency.

However, since the tiles of an image are not independent, encoding each one separately can be suboptimal and may result in a loss of global context. To address this, the current strategy involves adding the downscaled original image to the list of tiles, resizing it to match the resolution supported by the vision encoder. While this helps retain some of the overall context, it's not a perfect solution, as the reduced resolution of the original image makes it difficult to capture finer details and its resolution depends on the original image's resolution.

\paragraph{Can we do better than the image-splitting strategy?} An alternative to the image-splitting strategy and a promising direction for future research is to develop a vision encoder that can natively process images of varying resolutions, including very large ones, without changing the original aspect ratios, potentially incorporating a mechanism for handling long-context efficiently. This model could be trained efficiently using the Patch’n’Pack \citep{PatchNPack} strategyate a different number of visual tokens per image based on the original resolution, enabling the entire image to be encoded directly without the need to crop it into multiple sub-images.

\section{Training methods and datasets for VLMs}

\begin{figure}[ht]
\centering
\includegraphics[width=1.0\textwidth]{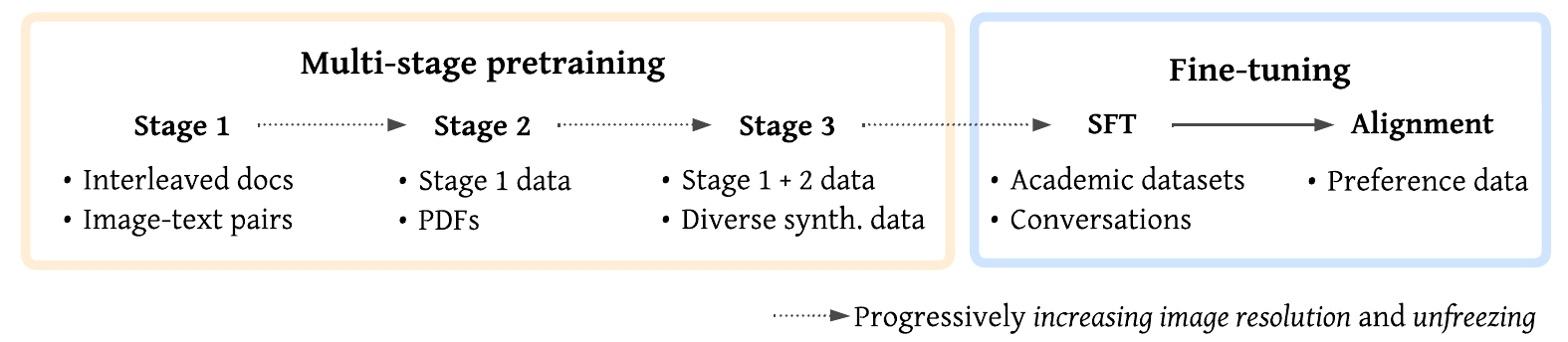}
\caption{The different stages of training and the types of datasets used.}
\label{fig:stages_training}
\end{figure}

Training VLMs typically occurs in multiple stages, primarily due to (a) the limited availability of high-quality data at scale, (b) memory constraints for efficient training, and (c) stability issues. During these stages, progressively higher-quality data is introduced, the maximum image resolution is gradually increased, and more model parts are unfrozen. Figure \ref{fig:stages_training} illustrates the key stages of training and the types of datasets used at each stage. As discussed in the previous section, the process begins with two unimodal pre-trained backbones: a language model and a vision encoder.

\begin{figure}[ht]
\centering
\includegraphics[width=1.0\textwidth]{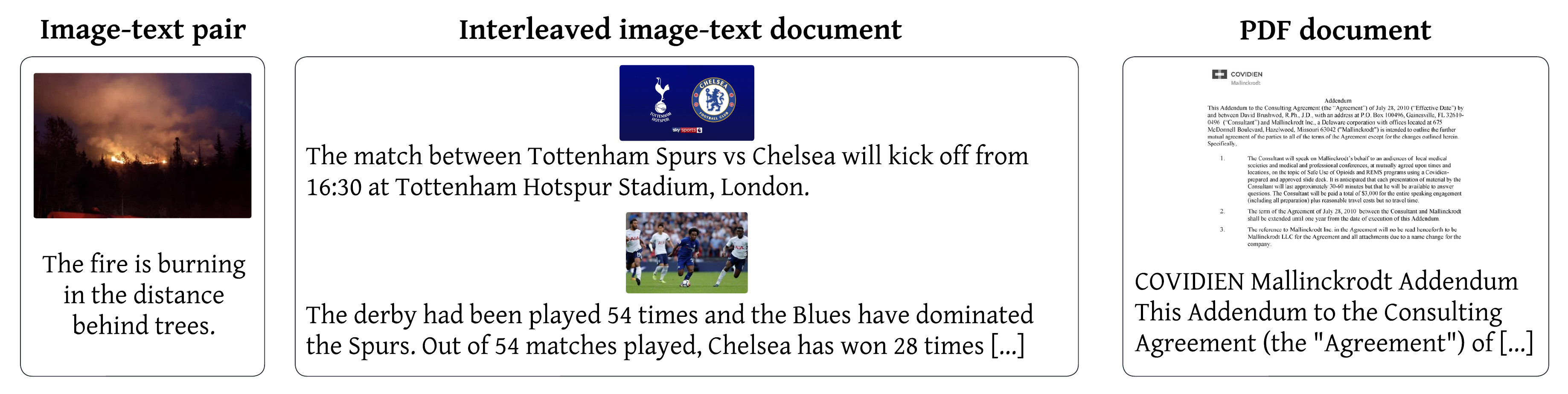}
\caption{Types of examples used during the pre-training of VLMs. (a) An image-text pair from LAION COCO, (b) an interleaved image-text document from OBELICS, (c) a PDF document from OCR-IDL.}
\label{fig:types_pretraining_datasets}
\end{figure}

\subsection{Multi-stage pre-training}\label{sec:multi_stage_pre_training}

The primary goal of pre-training is to align the backbone models and train the newly initialized parameters in the model. This is achieved using large-scale datasets to expose the VLM to a wide variety of examples to build extensive knowledge and improve robustness against out-of-domain data. To preserve the initial performance of the LLM, some models, like VILA \citep{VILA} and LLaVA-NeXT \citep{LLAVA-NeXT}, begin training by freezing the backbone models and focusing solely on the newly initialized parameters (the connector) until a satisfactory performance level is achieved. Afterward, the vision encoder and/or the language model can be gradually unfrozen. If instabilities arise, or if there's a need to enhance the model's expressivity while adding more regularization than full unfreezing, a LoRA \citep{LoRA} approach can be effective even during the pre-training phase \citep{Idefics2}.

To efficiently train on a large number of images, the image resolution is typically kept low at the start of training and gradually increased over time. Once the resolution is sufficiently high, datasets containing large images, such as PDFs, can be incorporated into the training data.

In the following paragraphs, we will discuss the various types of data typically used during this process. Examples of the most common ones are illustrated in Figure \ref{fig:types_pretraining_datasets}.

\paragraph{Image-text pairs}

Image-text pair datasets are generally created by crawling the web, downloading images, and extracting the corresponding alt-text from the original HTML files. Due to the ease of collecting these raw image-text pairs and their effectiveness in establishing strong alignment between images and text, many large-scale datasets have been created, such as LAION \citep{LAION-5B} with 5B images, COYO \citep{COYO-700M} with 700M images, and DataComp \citep{DataComp} with 12.8B images.

However, the alt-texts in these datasets are often noisy, ungrammatical, or too brief, making training challenging. Recent approaches have achieved better results by using synthetic re-captioning, where the same images from the original datasets are re-captioned using another model \citep{MM1, DallE3, Idefics2}. For example, LAION COCO \citep{LAION-COCO} re-captioned 600 million images from LAION-5B using an ensemble of BLIP \citep{BLIP} and two CLIP models \citep{CLIP}. Similarly, VeCap \citep{VeCap} combines the original alt-text with a synthetically generated caption from LLaVA \citep{LLaVA} to create a dataset of 300 million samples.

While these efforts have mainly focused on generating high-quality captions for given images, less attention has been paid to the initial selection of "good" images, which remains a promising area of research. This is important given the high proportion of web images that may not be useful for VLM training (e.g., logos, icons, portraits of non-public figures). Synth2 \citep{Synth2} addresses this by reversing the usual process, starting with LLM-generated captions and then using a Text-to-Image model to generate corresponding images. Furthermore, studies such as SNIP-Dedup \cite{SNIP-Dedup} and SemDeDup \citep{SemDeDup} have shown that by applying image deduplication, it is possible to train on just half of the LAION dataset with only a minimal reduction in performance compared to using the full dataset.

\paragraph{Interleaved image-text documents}

Training on interleaved image-text documents, also called web documents, was first introduced in Flamingo \citep{Flamingo} using the proprietary M3W dataset. OBELICS \citep{OBELICS} is an open-source dataset of interleaved image-text documents, containing 141 million documents and 353 million images. This dataset was constructed from HTML files obtained from Common Crawl dumps, which were carefully filtered. The resulting documents maintain the original linearity of images and texts as they appeared on the websites, while removing spam and ads.\newline
The authors highlight several advantages of using web documents in the training data mix: (a) it enhances in-context learning abilities, (b) it improves the model's ability to understand an arbitrary number of images interleaved with text, and (c) it exposes the model to a much wider distribution of texts than what is available in standard image-text pair datasets. This aligns with findings from MM1 \citep{MM1}, which showed that interleaved data is instrumental for few-shot and text-only performance. OBELICS has been used in the training of various VLMs, including MM1 \citep{MM1}, Idefics2 \citep{Idefics2}, and BLIP-3 \citep{BLIP-3}. Recently, the scale of these datasets has been significantly expanded, with MINT-1T \citep{MINT-1T} growing to 1T documents and 3.4B images, and OmniCorpus \citep{OmniCorpus} reaching 2.2B documents and 8.6B images.\newline
Model-based filtering on educational content, similar to the approach in Phi-3 \citep{Phi-3} and FineWeb-Edu \citep{FineWeb-Edu}, remains unexplored for these multimodal datasets and could likely offer significant improvements.

\paragraph{PDF documents}

Two primary datasets for PDF documents paired with their text transcriptions are OCR-IDL \citep{OCRIDL} and PDFA\footnote{\url{https://huggingface.co/datasets/pixparse/pdfa-eng-wds}}. OCR-IDL includes 26M pages of industry documents, while the English-only filtered version of PDFA contains 18M pages sourced from Common Crawl, offering greater diversity than OCR-IDL. Both datasets were created using OCR extraction tools to obtain corresponding texts and their locations within the documents, which can be linearized into a full document transcription. Idefics2 \citep{Idefics2} used these datasets directly during pre-training, an approach also adopted at scale in Llama 3-V \citep{Llama3} to enhance performance on document understanding tasks.

\paragraph{Synthetic data}

sh foundational skills such as (a) image captioning, (b) handling an arbitrary number of images interleaved with diverse texts, and (c) text transcription, all of which are essential for tackling more complex tasks. These datasets are abundant, as they are primarily built by crawling the web, ensuring a broad distribution of texts and images, enhancing robustness against rare examples.\newline
However, these datasets fall short in addressing many of the tasks that users typically require, such as document understanding or visual math reasoning, which are significantly more challenging. Relying on generalization or the limited examples in current fine-tuning datasets to master these tasks is not ideal.\newline
In the training of LLMs, synthetic data has proven to be highly effective \citep{Vicuna, phi-1, SyntheticDataReviewGoogle, Llama3}. Given the recent advancements in VLMs, which now solve many real-world examples with high accuracy, creating and training on large-scale synthetic datasets is a logical step. These datasets can be tailored to include examples that closely resemble the tasks users will likely request, making them more relevant than the data used in earlier training stages.\newline
The main categories of synthetic data that could be used are outlined below.

\begin{description}
\item \textbf{Image captioning} \hspace{0.15cm} The leading dataset for images paired with detailed captions is PixelProse \citep{PixelProse}. This dataset, built using images from CC12M \citep{CC12M}, CommonPool \citep{DataComp}, and RedCaps \citep{RedCaps}, contains captions generated by Gemini 1.0 Pro \citep{Gemini}. Despite being smaller in scale with 17M images, PixelProse offers richer descriptions and uses a stronger model for caption generation, making it an improvement over LAION COCO. Future improvements could include a more diverse, filtered, and deduplicated set of images, better models to reduce potential hallucinations in the generations, and various prompts for stylistic diversity. A similar dataset, ShareGPT-4o\footnote{\url{https://huggingface.co/datasets/OpenGVLab/ShareGPT-4o}}, re-captions images using GPT-4o to obtain 57K examples.

\item \textbf{Real-world visual question answering} \hspace{0.15cm} Datasets in this category contain QA pairs about real-world images, covering topics like identifying people or objects, understanding subtle scenes, counting, color identification, or spatial positioning. The leading dataset in this area is LNQA\footnote{\url{https://huggingface.co/datasets/vikhyatk/lnqa}}, with 300K images sourced from Localized Narratives \citep{LocalizedNarratives} and 1.5M QA pairs.

\item \textbf{Text reading in natural images} \hspace{0.15cm} In LLAvAR \citep{LLaVAR}, the authors use OCR tools to extract text from real-world images in the LAION-5B dataset \citep{LAION-5B}, resulting in 420K samples. Similar approaches are seen in MiniCPM-V \citep{MiniCPM-V} and Llama 3-V \citep{Llama3}. The key advantage of these datasets is their scalability and the unique distribution of text in natural images compared to PDF documents, which enhances the model’s ability to tackle tasks like TextVQA \citep{textvqa}.

\item \textbf{Text transcription} \hspace{0.15cm} The leading dataset for text transcription is PDFA, mentioned above. However, linearizing texts coherently from bounding boxes can be challenging, and math equations are often inaccurately transcribed or omitted, an area where models like Nougat \citep{Nougat} excel. Additionally, figures and tables are often poorly transcribed by OCR tools. A better strategy for text transcription would involve combining a traditional OCR tool, a document-specialized model like Nougat, and a robust VLM to judge, refine, and merge the outputs of these models.

\item \textbf{Document understanding} \hspace{0.15cm} Understanding documents from images is complex, making the generation of quality synthetic QA pairs challenging even for advanced VLMs. However, accurate text transcriptions from document images can be obtained with OCR tools, and text-only LLMs are performant at generating QA pairs from these transcriptions. This approach was used to create the dataset Docmatix, introduced in detail later in this paper, which includes 1.3M documents up to 4 pages long and 9.5M QA pairs. Enhancements could involve generating more diverse questions, such as summarizing a paragraph, and employing a strong VLM to filter out erroneous generated QA pairs.

\item \textbf{Chart understanding} \hspace{0.15cm} ChartGemma \citep{ChartGemma} uses Gemini 1.5 Flash \citep{Gemini1.5} to generate 160K QA pairs for chart analysis, covering a range of questions like summarizing insights, converting charts to Markdown tables, and assessing the validity of stated facts based on the chart.

\item \textbf{Table understanding} \hspace{0.15cm} A dataset for table understanding can be created by either using a strong VLM with table images taken from the web, or by synthetically generating tables with an LLM, rendering them to images, and generating QA pairs with the LLM. However, to our knowledge, there is currently no large-scale open-source synthetic dataset available for this task.

\item \textbf{Reasoning with chain-of-thought} \hspace{0.15cm} In Meteor \citep{Meteor}, the authors developed a proprietary dataset to enable a model to answer complex questions using a chain-of-thought strategy \citep{CoT}. They began by collecting challenging QA pairs from academic datasets, where the answers were provided without explanations. Then, they employed Claude 3 Haiku \citep{Claude3} to generate detailed and comprehensive rationales for these answers. These rationales were finally filtered by GPT-4V \citep{GPT4} to ensure quality, resulting in a final set of 1.1M question-rationale-answer triples.

\item \textbf{Visual mathematical reasoning} \hspace{0.15cm} Even the most advanced VLMs currently struggle with complex mathematical reasoning and geometry tasks. Generating synthetic data directly from a teacher model is problematic because the teacher often fails to provide correct answers. Instead, datasets like Geo170K \citep{G-llava-Geo170K} and MAVIS-Instruct \citep{MAVIS} are created by augmenting small and accurate academic mathematical datasets using an LLM. In AlphaGeometry \citep{OlympiadGeometryDeepmind}, the authors train a model exclusively on synthetically generated geometric problems, enabling it to solve olympiad-level challenges effectively.

\item \textbf{Converting web screenshots into HTML code} \hspace{0.15cm} To develop models capable of efficiently converting web screenshots into functional HTML code, WebSight \citep{WebSight} introduced a fully synthetic dataset comprising 2M pairs of HTML code and their corresponding screenshots. The HTML and TailWind CSS code were generated using DeepSeek-Coder \citep{DeepSeek-Coder}, merged into a single file, and then filtered and rendered to obtain the web screenshot. Instead of relying on a general LLM coder, further improvements could be achieved by using a specialist LLM fine-tuned specifically for HTML and CSS generation, enabling the creation of more diverse and visually appealing websites. In InternLM-XComposer-2.5 \citep{InternLM-XComposer-2.5}, in addition to the WebSight dataset, the authors built a proprietary dataset that includes HTML and CSS files from The Stack v2 \citep{StarCoder-TheStackv2} which were heavily filtered to remove external links and irrelevant content. This approach benefits from more diverse websites in the dataset, though it may introduce challenges with potentially noisy, lengthy, or difficult-to-learn examples.

\item \textbf{Locating objects in an image} \hspace{0.15cm} Determining the exact positions of objects within an image by generating bounding boxes around them is useful for various applications, such as enabling a VLM to navigate the web by selecting where to click based on positional output. In BLIP3-GROUNDING-50M \citep{BLIP-3}, large-scale grounding datasets are created by using a diverse set of images, where objects and their locations are identified using open-world image tagging and object detection models.

\end{description}

\subsection{Fine-tuning}

Similar to the approach commonly used with LLMs \citep{Llama2}, fine-tuning is typically done in two stages: supervised fine-tuning (SFT) followed by an alignment phase.

\paragraph{Which datasets should be used for the SFT?}

The literature offers many high-quality datasets containing diverse images and covering a wide range of tasks. They are often annotated by humans, ensuring accurate QA pairs. Although most of them are relatively small individually, when combined, they provide a sufficient number of examples for an effective SFT.\newline
Inspired by previous work on LLMs \citep{flan, T0}, InstructBLIP \citep{InstructBLIP} and M3IT \citep{m3it} were among the first to introduce curated mixtures of academic datasets for fine-tuning VLMs. Building on these efforts, The Cauldron \citep{Idefics2} introduced a collection of 50 high-quality datasets covering a broad range of tasks, including general visual question answering, counting, captioning, text transcription, document understanding, chart/figure analysis, table understanding, visual reasoning, geometry, spotting differences between two images, and converting screenshots into functional code. Each dataset in this compilation is formatted into a standardized question/answer format, and when multiple QA pairs exist per image, they are combined into a multi-turn conversation. However, a drawback of academic datasets is that their answers tend to be concise, which may lead the model to generate similarly brief responses, which are often less preferred by users. A potential solution is to use an LLM to expand and rephrase the answers, as in M3IT \citep{m3it} and Llava 3-V \citep{Llama3}.

\paragraph{Alignment phase} 

There are several reasons to include an alignment stage following supervised fine-tuning. The first objective is to align the model’s output with human preferences, making it more intuitive and better at following complex instructions. Additionally, as demonstrated in RLHF-V \citep{RLHF-V}, this stage effectively reduces hallucinations, where the model might describe objects or details not actually present in the image. It also enhances model safety by minimizing the risk of generating harmful content. It also may further improve overall model performance.\newline
RLAIF-V \citep{RLAIF-V} provides a dataset of 80K preference pairs, used in the training of MiniCPM-V 2.5 \citep{MiniCPM-V}. VLFeedback \citep{VLFeedback} offers 380K comparison pairs, where model responses sampled from 12 VLMs are ranked by GPT-4V \citep{GPT4}. Similarly, SPA-VL \citep{SPA-VL} generates 100K preference pairs through a comparable approach. DPO \citep{DPO} is then commonly applied to these datasets during the alignment phase.

\section{Challenges in evaluating VLMs}

\subsection{Open-ended and multiple-choice benchmarks}

The earliest and most popular multimodal benchmarks, such as VQAv2 \citep{VQAv2}, OKVQA \citep{okvqa}, TextVQA \citep{textvqa}, and COCO Captioning \citep{coco}, are mainly open-ended. These benchmarks rely on specific ground-truth answers for each question, so even minor variations in the model’s responses can lead to a score marked as incorrect.\newline
This method of evaluation tends to favor models that produce answers closely aligned with the benchmark's expected format or writing style. For example, VQAv2, which assesses general real-world image understanding, typically expects short answers, often just one or two words. Even when the evaluation prompt clearly specifies this format, models like Gemini 1.0 Ultra \citep{Gemini} and GPT-4V \citep{GPT4} achieve scores of 77.8 and 77.2, respectively. These scores are notably lower than those of much smaller models that include a small portion of VQAv2 in their fine-tuning data: MM1-3B-Chat \citep{MM1} reaches 82.0, and moondream2 achieves 79.4 with only 1.9B parameters. This discrepancy highlights the challenge of evaluating different models without letting the benchmark's template influence the results.\newline
One potential way to mitigate this bias is to perform few-shot evaluations, although this approach is less effective than training on the benchmark training set, and is not currently used for evaluating instruct models.\newline
However, the level of ambiguity in these evaluations can vary by benchmark. For instance, TextVQA and DocVQA \citep{DocVQA} require the model to read and extract text directly from an image without rephrasing it, which reduces ambiguity. In MathVista \citep{mathvista}, where answers are always numerical, each question is paired with specific instructions, such as indicating whether the answer should be an integer or a float rounded to two decimal places.\newline
Recently proposed, the LAVE metric \citep{LAVE} consists of asking an LLM to evaluate whether the response generated by the VLM is correct, given the ground truth and the specific question, thereby reducing the template problem.\newline
Another way to reduce ambiguity is to use benchmarks that include multiple-choice questions (MCQs), where the model selects the correct option by choosing the corresponding letter. Many recent benchmarks have adopted this approach, such as MMMU \citep{MMMU}, MMStar \citep{MMStar}, and MMBench \citep{MMBench}.

\subsection{Challenges in model evaluation during the pre-training stage}

There is a significant discrepancy between the performance of VLMs at the pre-training stage versus after fine-tuning. For instance, Idefics2-base \citep{Idefics2} scores 57.9 on TextVQA \citep{textvqa} using 8 in-context examples and less than 55 on DocVQA \citep{DocVQA} during pre-training. However, after fine-tuning, it achieves 70.4 on TextVQA and 67.3 on DocVQA in a zero-shot setting, without employing the image-splitting strategy. As noted earlier, these open-ended tasks are less influenced by the specific template expected by the benchmark.

One reason for this gap is that the model only starts learning the specific task of visual question answering (beyond just image captioning or text transcription) during the fine-tuning stage—unless a third pre-training stage is conducted using large synthetic VQA datasets, as described in Figure \ref{fig:stages_training}, which offer examples more aligned with the ones present in benchmarks.

When instruction data is omitted during pre-training, more complex tasks like document understanding may perform poorly, and the impact of development choices in the VLM may only become evident after fine-tuning, leading to a delayed feedback loop. This delay can make pre-training ablations misleading. For example, in Idefics2, the authors found no noticeable improvements during pre-training when using 128 visual tokens instead of 64 with their architecture. While this held true for most tasks, the benefit of using more visual tokens per image became apparent in OCR tasks after fine-tuning with the image-splitting strategy. Therefore, to obtain more accurate insights during pre-training ablations, we recommend incorporating instruction data into the data mixture.

\subsection{Risk of contamination and overoptimization in some benchmarks}

Some benchmarks are derived from the validation or test sets of existing academic datasets. For instance, MathVista \citep{mathvista}, a leading benchmark for evaluating reasoning and math capabilities, shows signs of potential contamination. We found that at least 6.6\% of the questions include images from the training sets of academic datasets often used in supervised fine-tuning, and 2.2\% feature both an image and a question that is identical or highly similar.

Additionally, this benchmark often includes questions that are especially difficult to answer unless the model has encountered them during training. For example, we find that at least 6.1\% of the questions in MathVista ask variations of the question, \texttt{"What is the age gap between these two people in the image?"}. Variants of this question are also abundant on KVQA \citep{KVQA}. Therefore, models incorporating  in their fine-tuning data will have an advantage for MathVista.

Ultimately, benchmarks should be used to measure model performance, not as a training objective. Fine-tuning on similar examples can boost scores, but it provides little evidence for the model's ability to generalize to real-world scenarios. Thus, we encourage researchers to exclude images used in the benchmarks they evaluate from their supervised fine-tuning data.

\section{Idefics3: adapting Llama 3 to multimodality}

In this section, we detail the construction of Idefics3, a VLM based on Llama 3.1 \citep{Llama3} and SigLIP-SO400M \citep{SigLIP}. First, we begin by preparing the dataset used for training.

\subsection{Dataset preparation}

Our approach mainly takes the datasets used in the training of Idefics2 \citep{Idefics2} while also adding complementary datasets for supervised fine-tuning to expand the range of tasks covered. These datasets are detailed below.

\subsubsection{Extending The Cauldron}

As previously mentioned, The Cauldron \citep{Idefics2} is a collection of 50 high-quality datasets from existing literature. We have expanded this collection by adding 6 more datasets: Cord-v2\footnote{\url{https://huggingface.co/datasets/naver-clova-ix/cord-v2}} for training models to output information in JSON format, LNQA for large-scale real-world visual question answering, ShareGPT-4o and IIW-400 \citep{IIW-400} for generating detailed captions, Geo170K \citep{G-llava-Geo170K} for tasks involving geometry, and Docmatix for document understanding.

In Table \ref{table:mixture_sft}, we present the statistics of the datasets included in The Cauldron and the text-only instruction datasets used for the supervised fine-tuning. For each dataset, we give the number of different images it contains, the number of question-answer pairs, the total number of tokens for the answers in the question-answer pairs, and the selected percentage of answer tokens it represents in our final mixture after upsampling or downsampling.

\begin{figure}[ht]
\centering
\includegraphics[width=1.0\textwidth]{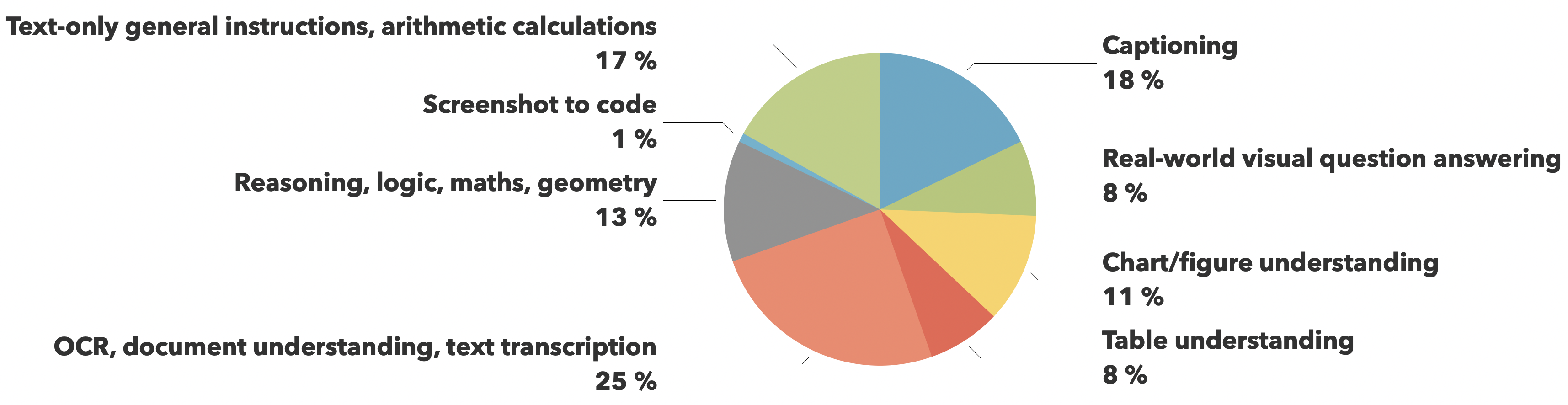}
\label{fig:mixture_the_cauldron}
\end{figure}

\vspace{-0.5cm}

\begin{longtable}{lcccc}
\toprule
\textbf{Dataset} & \textbf{\makecell{\# images}} & \textbf{\# QA pairs} & \textbf{\# tokens} & \textbf{\% mix}\\ \midrule
 &  &  &  & \\
\multicolumn{5}{l}{\textit{Captioning}} \\
ShareGPT-4o \footnote{\url{https://huggingface.co/datasets/OpenGVLab/ShareGPT-4o}} & 57,259 & 57,259 & 39,696,010 & 13.03\% \\
LNarratives \citep{LocalizedNarratives} & 507,444 & 507,444 & 21,328,731 & 1.40\% \\
TextCaps \citep{textcaps} & 21,953 & 21,953 & 389,658 & 1.28\% \\
VisText \citep{VisText} & 7,057 & 9,969 & 1,245,485 & 1.23\% \\
IIW-400 \citep{IIW-400} & 400 & 400 & 103,024 & 0.68\% \\
Screen2Words \citep{screen2words} & 15,730 & 15,743 & 143,103 & 0.23\% \\
 &  &  &  & \\
\multicolumn{5}{l}{\textit{Real-world visual question answering}} \\
LNQA \footnote{\url{https://huggingface.co/datasets/vikhyatk/lnqa}} & 302,780 & 1,520,942 & 21,107,241 &3.46\% \\
VQAv2 \citep{VQAv2} & 82,772 & 443,757 & 1,595,929 &2.10\% \\
COCO-QA \citep{CocoQA} & 46,287 & 78,736 & 286,982 & 0.94\% \\
Visual7W \citep{Visual7w} & 14,366 & 69,817 & 279,268 & 0.92\% \\
OK-VQA \citep{okvqa} & 8,998 & 9,009 & 38,853 & 0.26\% \\
VSR \citep{VSR} & 2,157 & 3,354 & 10,062 & 0.13\% \\
 &  &  &  & \\
\multicolumn{5}{l}{\textit{OCR, document understanding, text transcription}} \\
Docmatix\footnote{\url{https://huggingface.co/datasets/HuggingFaceM4/Docmatix}} (ours) & 1,273,215 & 9,488,888 & 392,302,612 & 10.31\% \\
RenderedText\footnote{\url{https://huggingface.co/datasets/wendlerc/RenderedText}} & 999,000 & 999,000 & 27,207,774 & 7.15\% \\
DocVQA \citep{DocVQA} & 10,189 & 39,463 & 337,829 & 2.22\% \\
TextVQA \citep{textvqa} & 21,953 & 34,602 & 181,918 & 1.19\% \\
Cord-v2 \footnote{\url{https://huggingface.co/datasets/naver-clova-ix/cord-v2}} & 800 & 800 & 178,388 & 1.17\% \\
ST-VQA \citep{STVQA} & 17,247 & 23,121 & 127,846 & 0.84\% \\
OCR-VQA \citep{OCR-VQA} & 165,746 & 801,579 & 6,073,824 & 0.60\% \\
VisualMRC \citep{VisualMRC} & 3,027 & 11,988 & 168,828 & 0.55\% \\
IAM \citep{IAM} & 5,663 & 5,663 & 144,216 & 0.47\% \\
InfoVQA \citep{InfographicVQA} & 2,118 & 10,074 & 61,048 & 0.40\% \\
Diagram image-to-text\footnote{\url{https://huggingface.co/datasets/Kamizuru00/diagram_image_to_text}} & 300 & 300 & 22,196 & 0.07\% \\
 &  &  &  & \\
\multicolumn{5}{l}{\textit{Chart/figure understanding}} \\
Chart2Text \citep{Chart2Text} & 26,985 & 30,242 & 2,852,827 & 4.38\% \\
DVQA \citep{DVQA} & 200,000 & 2,325,316 & 8,346,234 & 4.27\% \\
ChartQA \citep{ChartQA} & 18,271 & 28,299 & 185,835 & 1.90\% \\
PlotQA \citep{PlotQA} & 157,070 & 20,249,479 & 8478299.278 & 0.65\% \\
FigureQA \citep{FigureQA} & 100,000 & 1,327,368 & 3,982,104 & 0.61\% \\
MapQA \citep{MapQA} & 37,417 & 483,416 & 6,470,485 & 0.33\% \\
 &  &  &  & \\
\multicolumn{5}{l}{\textit{Table understanding}} \\
TabMWP \citep{TabMWP} & 22,729 & 23,059 & 1,948,166 & 1.60\% \\
TAT-QA \citep{TAT-QA} & 2,199 & 13,215 & 283,776 & 1.40\% \\
HiTab \citep{Hitab} & 2,500 & 7,782 & 351,299 & 1.15\% \\
MultiHiertt \citep{Multihiertt} & 7,619 & 7,830 & 267,615 & 0.88\% \\
FinQA \citep{FinQA} & 5,276 & 6,251 & 242,561 & 0.64\% \\
WikiSQL \citep{WikiSQL} & 74,989 & 86,202 & 9,680,673 & 0.64\% \\
SQA \citep{SQA} & 8,514 & 34,141 & 1,894,824 & 0.62\% \\
TQA \citep{TQA} & 1,496 & 6,501 & 26,004 & 0.34\% \\
WTQ \citep{WTQ} & 38,246 & 44,096 & 6,677,013 & 0.33\% \\
 &  &  &  & \\
\multicolumn{5}{l}{\textit{Reasoning, logic, maths, geometry}} \\
Geo170K \citep{G-llava-Geo170K} & 9,067 & 177,457 & 17,971,088 & 2.95\% \\
GeomVerse \citep{GeomVerse} & 9,303 & 9,339 & 2,489,459 & 2.45\% \\
CLEVR-Math \citep{CLEVR-Math} & 70,000 & 788,650 & 3,184,656 & 2.09\% \\
CLEVR \citep{CLEVR} & 70,000 & 699,989 & 2,396,781 & 0.79\% \\
A-OKVQA \citep{A-OKVQA} & 16,539 & 17,056 & 236,492 & 0.78\% \\
IconQA \citep{IconQA} & 27,315 & 29,859 & 112,969 & 0.74\% \\
AI2D \citep{AI2D} & 3,099 & 9,708 & 38,832 & 0.51\% \\
NLVR2 \citep{NLVR2} & 50,426 & 86,373 & 259,119 & 0.43\% \\
RAVEN \citep{RAVEN} & 42,000 & 42,000 & 105,081 & 0.43\% \\
TallyQA \citep{TallyQA} & 98,680 & 183,986 & 738,254 & 0.36\% \\
Spot the diff \citep{SpotTheDiff} & 8,566 & 9,524 & 221,477 & 0.36\% \\
GSD \citep{MIMIC-IT-General-Scene-Difference} & 70,939 & 141,869 & 4,637,229 & 0.30\% \\
ScienceQA \citep{ScienceQA} & 4,985 & 6,218 & 24,872 & 0.16\% \\
Inter-GPs \citep{Inter-GPS} & 1,451 & 2,101 & 8,404 & 0.11\% \\
HatefulMemes \citep{hatefulmeme} & 8,500 & 8,500 & 25,500 & 0.08\% \\
 &  &  &  & \\
\multicolumn{5}{l}{\textit{Screenshot to code}} \\
WebSight \citep{WebSight} & 500,000 & 500,000 & 276,743,299 & 0.91\% \\
DaTikz \citep{DaTikz} & 47,974 & 48,296 & 59,556,252 & 0.02\% \\
 &  &  &  & \\
 \midrule  \\
\multicolumn{5}{l}{\textit{Text-only general instructions, math problems, arithmetic calculations}} \\ 
OpenHermes-2.5 \citep{OpenHermes} & 0 & 1,006,223 & 248,553,747 & 8.16\% \\
MetaMathQA \citep{MetaMathQA} & 0 & 395,000 & 74,328,255 & 2.44\% \\
AtlasMathSets\footnote{\url{https://huggingface.co/datasets/AtlasUnified/atlas-math-sets}} & 0 & 17,807,579 & 455,411,624 & 2.24\% \\
MathInstruct \citep{MathInstruct} & 0 & 261,781 & 45,393,559 & 1.49\% \\
OrcaMath \citep{Orca-Math} & 0 & 200,031 & 63,780,702 & 1.05\% \\
Goat \citep{Goat} & 0 & 1,746,300 & 167,695,693 & 0.55\% \\
LIMA \citep{LIMA} & 0 & 1,052 & 633,867 & 0.52\% \\
Dolly \citep{Dolly} & 0 & 14,972 & 1,329,999 & 0.44\% \\
CamelAIMath \citep{CamelAIMath} & 0 & 49,744 & 21,873,629 & 0.04\% \\
\bottomrule
\vspace{-0.5em} \\
\caption{The statistics of datasets used for instruction fine-tuning. \# tokens is the total number of tokens for each dataset for the answers only. \% mix is our selected percentage of answer tokens for each dataset in the final mixture.}
\label{table:mixture_sft}
\end{longtable}

\subsubsection{Enhancing document understanding capabilities with Docmatix}

Document understanding is a critical business application for VLMs. Yet, only a few open-source datasets are available for boosting the performance of models in this area, and they typically include only a limited number of examples. For instance, DocVQA \citep{DocVQA} offers 10K images and 40K QA pairs, InfographicVQA \citep{InfographicVQA} contains 2K images and 10K QA pairs, and VisualMRC \citep{VisualMRC} provides 3K images and 12K QA pairs.

Moreover, generating high-quality synthetic data for this task is relatively straightforward if we reframe the problem as one of LLM-based data generation rather than relying solely on VLMs. Standard OCR tools can accurately extract text from PDF documents, and an LLM can then be used to generate QA pairs based on this text.\newline
These motivations lead us to build a large-scale document understanding dataset.

\begin{figure}[ht]
\centering
\includegraphics[width=0.9\textwidth]{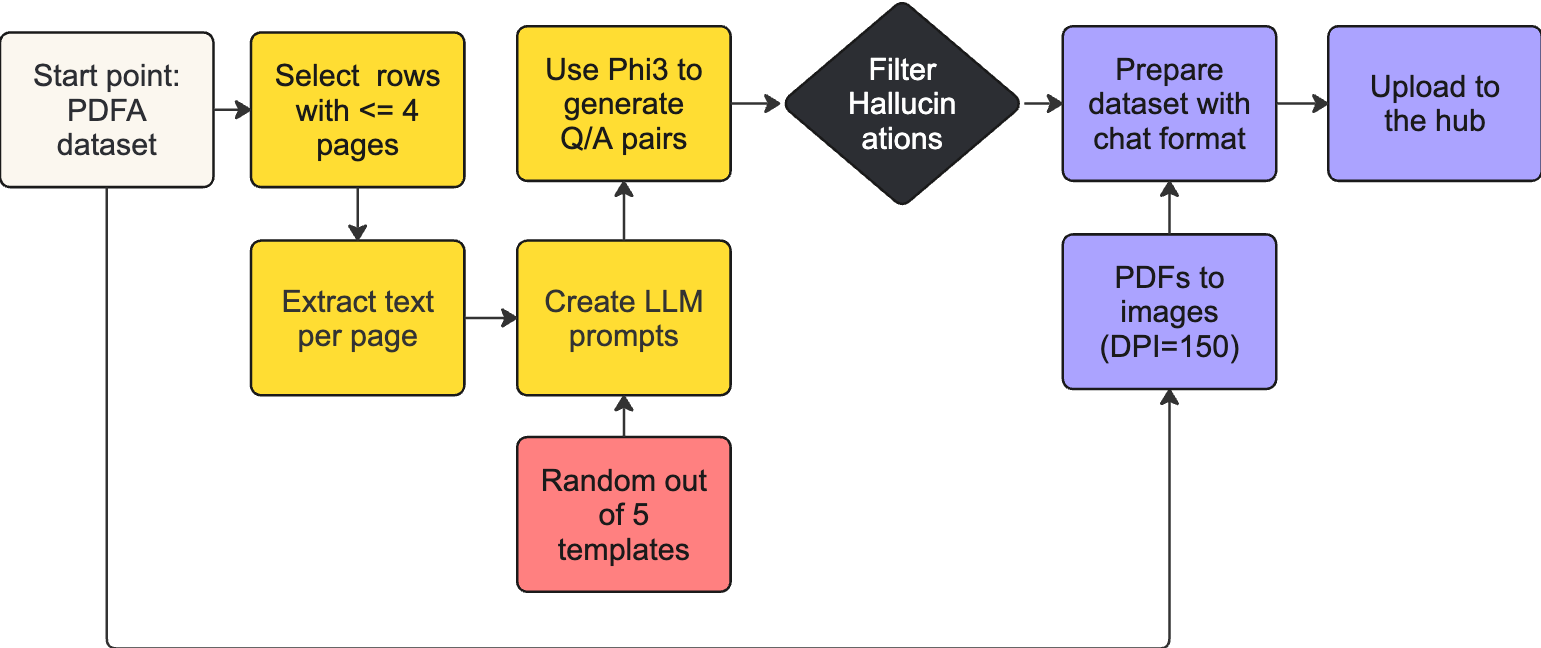}
\caption{Overview of the pipeline used for the creation of Docmatix.}
\label{fig:pipeline_docmatix}
\end{figure}

We begin with the text transcriptions from the English PDFA dataset and use Phi-3-small \citep{Phi-3} to generate QA pairs. To ensure diverse outputs, we employ five different prompts. To maintain dataset quality, we filter the results, discarding 15\% of QA pairs flagged as incorrect. This is done by using regular expressions to detect code and removing answers containing the keyword "unanswerable." Figure \ref{fig:pipeline_docmatix} shows an overview of our dataset creation pipeline.

The resulting dataset, Docmatix \footnote{\url{https://huggingface.co/datasets/HuggingFaceM4/Docmatix}}, includes 2.4M images and 9.5M QA pairs derived from 1.3M PDF documents, representing a 240-fold increase in scale compared to previous open datasets.

\setlength\intextsep{5pt}
\begin{wraptable}{r}{8cm} 
  \centering
  \scalebox{0.8}{
    \begin{minipage}{0.7\textwidth}
    \centering
      \begin{tabular}{ccc}
        \toprule
        \textbf{Model / Training data} & \textbf{Model size} & \textbf{DocVQA (ANLS)} \\
        \midrule
        Florence-2 / DocVQA & 700M & 60.1 \\
        Florence-2 / Docmatix & 700M & 71.4 \\
        Idefics2 / General mixture & 8B & 74.0 \\
        \bottomrule
      \end{tabular}
      \caption{Ablation on the importance of Docmatix to improve the performance on document understanding tasks.}
      \label{tab:ablations_docmatix}
    \end{minipage}
  }
\end{wraptable}

To assess Docmatix's effectiveness, we conduct ablation studies using the Florence-2 \citep{Florence-2} model. We train two versions of the model: one trained over multiple epochs on the DocVQA dataset, and another trained for a single epoch on a subset of Docmatix (20\% of images and 4\% of QA pairs), followed by an epoch on DocVQA to ensure proper format for evaluation. The results, shown in Figure \ref{tab:ablations_docmatix}, are significant: training on this small portion of Docmatix leads to a nearly 20\% relative improvement. Additionally, the specialist 0.7B Florence-2 model performs only 5\% worse than the much larger 8B Idefics2 \citep{Idefics2} generalist model.

Since Docmatix was made publicly available prior to this paper, it has already been used to enhance the performance of the moondream2 model\footnote{\url{https://huggingface.co/vikhyatk/moondream2}}, which achieved a 103\% improvement on DocVQA compared to its previous version.

\subsection{Building Idefics3}

\subsubsection{Architecture and training methods}

Following Idefics2 \citep{Idefics2}, we use SigLIP-SO400M \citep{SigLIP} for the vision encoder, and swap the language model for Llama 3.1 instruct \citep{Llama3}, as it significantly outperforms Mistral-7B \citep{Mistral7B}.

For the connector between these backbones, Idefics2 uses a perceiver resampler to encode each image up to 980x980 pixels into 64 visual tokens. With Idefics3, we aim to enhance OCR capabilities. To address the bottleneck for OCR tasks of having too few visual tokens per image, we replace the perceiver resampler with a simple pixel shuffle strategy. This method, as in InternVL-1.5 \citep{InternVL}, acts as a pooling technique that reduces the number of image hidden states by a factor of 4, encoding each image up to 364x364 pixels into 169 visual tokens.

During both training and inference, we follow the image-splitting strategy, where the original image is divided into a matrix of tiles of 364x364 pixels. The number of rows and columns in this matrix depends on the resolution of the original image. The vision encoder processes each tile separately, resulting in a sequence of visual tokens.\newline
However, because images have a 2D structure, and the number of rows and columns in the tile matrix varies for each image, linearizing these visual tokens into a single sequence can cause the model to lose the information about the original arrangement of tiles, making it difficult to reconstruct their positions accurately. To address this issue, we follow the common practice of inserting a text token `\textbackslash n` after each row of tiles, and of appending the original image, downscaled to 364x364 pixels, to the sequence of tiles to provide the model with the complete image in its entirety \citep{SPHINX, Internlm-xcomposer2-4khd}. Additionally, as in mPLUG-DocOwl-1.5 \citep{mPLUG-DocOwl-1.5}, we prepend each tile with the textual tokens `<row\_$x$\_col\_$y$>`, where $x$ and $y$ indicate the tile's position in the matrix.

\vspace{0.5cm}

\begin{table}[!ht]
\small
\centering
\begin{tabular}{c|cccc}
\toprule
& \multicolumn{3}{c}{\textbf{Pre-training}} & \textbf{SFT} \\
& \textit{Stage 1} & \textit{Stage 2} & \textit{Stage 3} & \\ \midrule
\textit{Number of steps} & 1000 & 3000 & 1500 & 5000 \\ \midrule
\textit{Learning rate (max, min)} & (10$^{-4}$, 10$^{-4}$) & (10$^{-4}$, 10$^{-4}$) & (10$^{-4}$, 0) & (5x10$^{-5}$, 0) \\ \midrule
\textit{Batch size} & \multicolumn{4}{c}{1024} \\ \midrule
\textit{Sequence length} & \multicolumn{4}{c}{10K} \\ \midrule
\textit{Max image resolution} & \makecell{364² \\ 364² $\rightarrow$ 728² \\ 728² $\rightarrow$ 1092² \\ 1092² $\rightarrow$ 1456² \\ 1456² $\rightarrow$ 1820²} & 1820² & 1820² & 1820² \\ \midrule
\textit{Backbones training} & Frozen & LoRA & LoRA & LoRA \\ \midrule
\textit{Data} & \makecell{• OBELICS \\ • LAION COCO} & \makecell{• OBELICS \\ • LAION COCO \\ • PDFA} & \makecell{• PDFA \\ • Docmatix \\ • Websight \\ • LNQA \\ • PixelProse \\ • ChartGemma} & • The Cauldron \\
\bottomrule
\end{tabular}
\vspace{0.7em}
\caption{The different training stages of Idefics3, along with the parameters and datasets used.}
\label{table:training_details_idefics3}
\end{table}

Details of the Idefics3 training process are summarized in Table \ref{table:training_details_idefics3}. The training involves three stages of pre-training followed by supervised fine-tuning.

In the first pre-training stage, the model's backbones remain frozen to preserve their performance while learning the newly initialized parameters. We gradually increase the maximum image resolution from 364² to 1820². From the second stage onward, we efficiently train the backbones using DoRA \citep{DoRA}, a variant of LoRA \citep{LoRA}, and introduce larger images into the training data. The final pre-training stage focuses on training with large synthetic datasets.

During the supervised fine-tuning phase, we apply NEFTune noise \citep{NEFTune} to the inputs and calculate the loss only on the answer tokens. The learning rate is kept constant during the first two pre-training stages but is linearly decayed to zero during the final pre-training stage and supervised fine-tuning. The entire training process, including restarts, is completed in 5 days on 32 H100 nodes.

\paragraph{Opportunities for improvement}

There are several straightforward opportunities for improvement. First, although we did not encounter instabilities when fully unfreezing the backbones, we opt for a LoRA approach to enhance training efficiency. However, we believe that carefully executed full unfreezing can lead to better performance. Additionally, during the first two pre-training stages, the loss function is far from converging, but we move to the next stage to reduce computational costs. In stage 3 of pre-training, only a fraction of the examples available in the chosen datasets are used, again to reduce computational demands. Further significant improvements can be achieved by creating and incorporating the synthetic datasets mentioned in Section \ref{sec:multi_stage_pre_training} into the stage 3 data mixture.

\subsubsection{Evaluation}

We evaluate Idefics3 on commonly adopted and challenging benchmarks: MMMU \citep{MMMU} for multidiscipline college-level problems, MathVista \citep{mathvista} for visual mathematical reasoning, MMStar \citep{MMStar} for general image understanding, DocVQA \citep{DocVQA} for document understanding, and TextVQA \citep{textvqa} for text reading on natural images. For Idefics3, we evaluate the benchmarks by resizing all images so that the longest side is 4x364 pixels. The exception is DocVQA, which has larger images, where we resize them to 5x364, matching the maximum resolution used during training. For Idefics2-70B, we resize the longest side of all images to 1960 pixels, the maximum resolution seen during its training. The prompts used for the evaluations are provided in Section \ref{sec:prompts_evaluation}.

\begin{figure}[ht]
\centering
\includegraphics[width=1.0\textwidth]{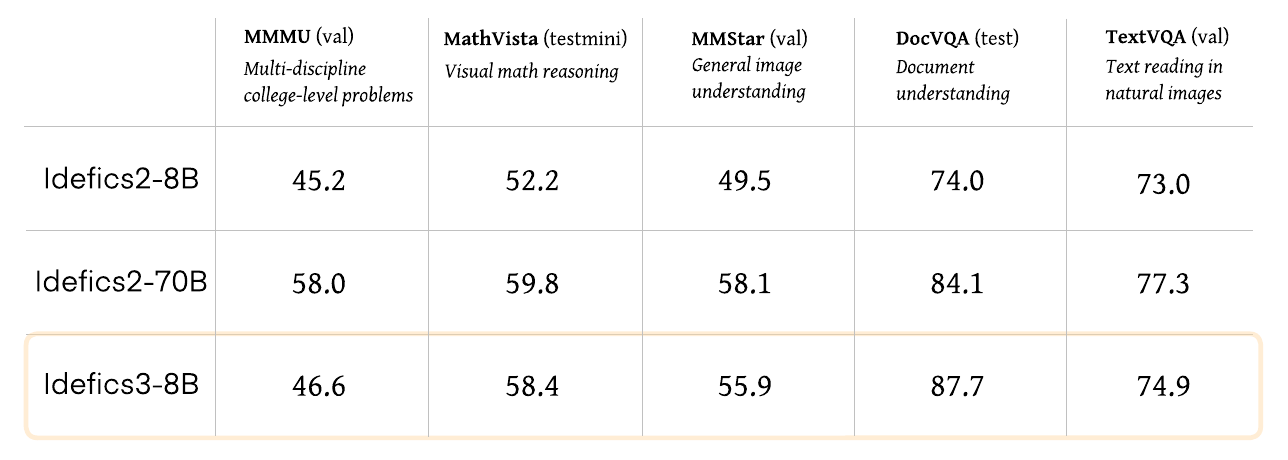}
\caption{Performance of Idefics3 against Idefics2-8B and Idefics2-70B. The evaluations are done in zero shot and without any chain-of-thought prompting.\\
\textit{(Benchmark, Split, Metric): (MMMU, val, MMMU score), (MathVista, testmini, MMMU score), (MMStar, val, acc.), (DocVQA, test, ANLS score), (TextVQA, val, VQA acc.).}}
\label{fig:performance_idefics3}
\end{figure}

Figure \ref{fig:performance_idefics3} presents the results of Idefics3 against Idefics2-70B and Idefics2-8B \citep{Idefics2}. The detailed performance of Idefics3 across each category of MMMU is present in Table \ref{table:detail_performance_mmmu}. Compared to Idefics2, Idefics3 benefits from having more visual tokens per image, a third stage of pre-training on large high-quality synthetic datasets, and an improved language model backbone. Despite being trained less extensively during the first two pre-training stages, these enhancements led to significant improvements across various tasks, particularly in document understanding tasks, with a boost of 13.7 points on DocVQA. However, the large gap of 11.4 points between Idefics2-70B and Idefics3-8B on MMMU indicates that scale is necessary for this benchmark to encapsulate sufficient knowledge into the model's weights.

\newgeometry{left=1cm,top=0.5cm,right=1cm,bottom=0.5cm}

\begin{figure}[hbtp]
\centering
\includegraphics[width=0.85\textwidth]{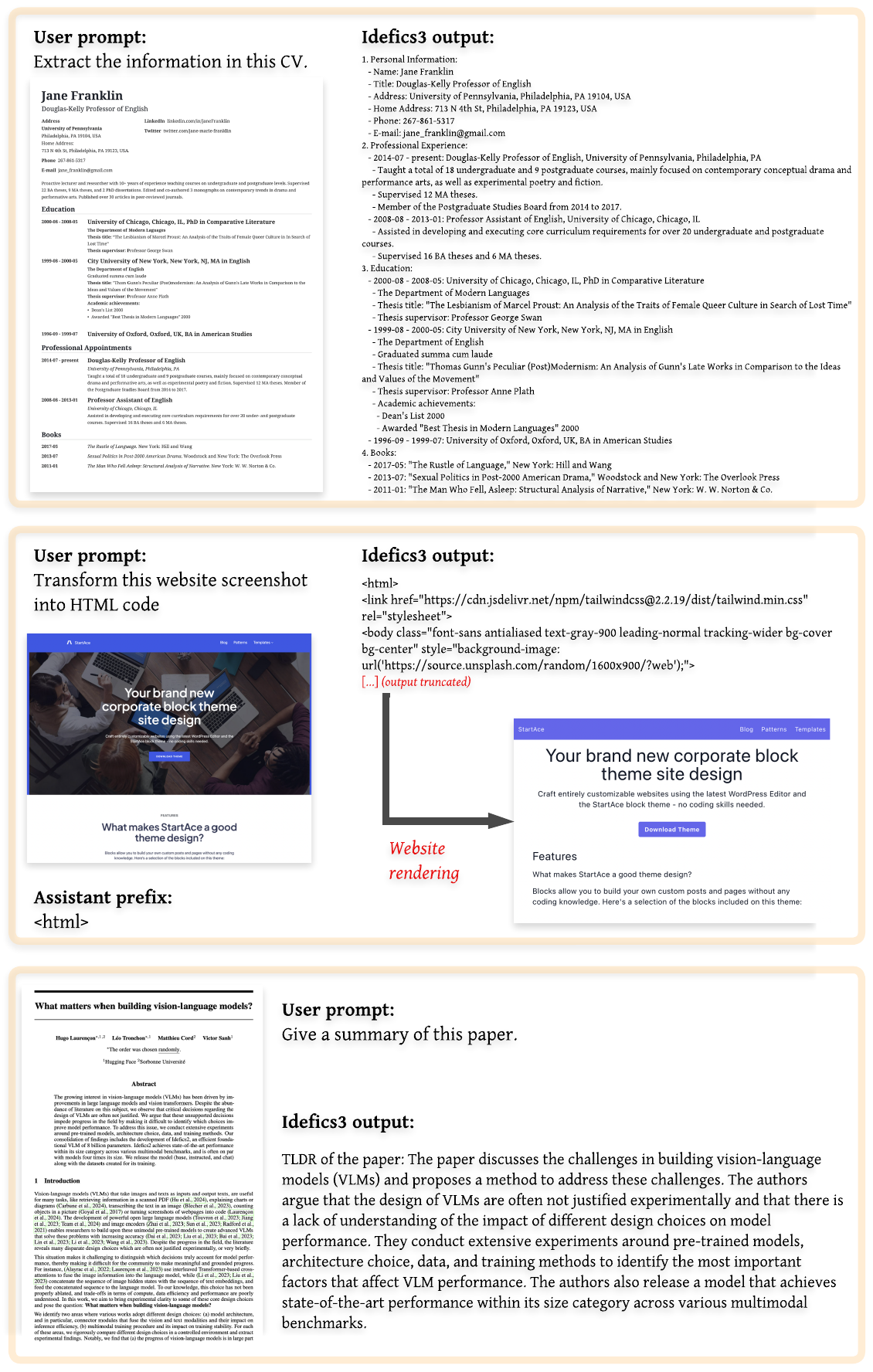}
\caption{Generation of Idefics3 on various examples.}
\label{fig:wow_examples_idefics3}
\end{figure}

\restoregeometry

\paragraph{Qualitative evaluation} In Figure \ref{fig:wow_examples_idefics3}, we present examples where Idefics3 is able to accurately extract information from a CV, generate HTML code to recreate a website from a screenshot, and summarize a research paper given a screenshot. However, since the model was mainly trained on short answers during supervised fine-tuning, and did not benefit from an alignment phase, we observe that it can sometimes struggle to follow instructions for more challenging prompts. Nonetheless, we find that adding a brief prefix to the assistant's response allows the user to easily shape the generated output as desired.

\section{Conclusion}

In this paper, we provided a comprehensive tutorial on building vision-language models (VLMs), emphasizing the importance of architecture, data, and training methods in the development pipeline. Through an in-depth analysis of current state-of-the-art approaches, we highlighted the strengths and weaknesses of various design choices and suggested potential research directions for improving the models. We then detailed the practical steps taken to build Idefics3-8B, a VLM that demonstrates significant improvements in document understanding tasks, particularly through the use of the introduced Docmatix dataset. By releasing both the model and the datasets, we aim to contribute to develop the next generation of responsible and open VLMs.

\section*{Acknowledgement}

We thank Amy Sartran for helping to integrate Idefics3 into the Transformers library \citep{huggingfaces-transformers}, Merve Noyan for building a demo for the model, and Leandro von Werra for reviewing this paper.

\newpage

\nocite{*}
\bibliographystyle{chicago}
\bibliography{sections/references_bib}

\newpage

\appendix

\section{Appendix}

\subsection{Evaluation of Idefics3}

\subsubsection{Prompts used for the evaluation}\label{sec:prompts_evaluation}

We evaluate MMStar \citep{MMStar} using our default template for multiple-choice questions, as seen during supervised fine-tuning:

\begin{tcolorbox}
Question: \{question\}\newline
Choices:\newline
A. \{choice\_a\}\newline
B. \{choice\_b\}\newline
C. \{choice\_c\}\newline
D. \{choice\_d\}\newline
...\newline
Answer with the letter.
\end{tcolorbox}

We evaluate MMMU \citep{MMMU} and MathVista \citep{mathvista} using the VLMEvalKit \citep{VLMEvalKit} library. For the multiple-choice questions in these benchmarks, we also use our default template.

For TextVQA \citep{textvqa} and DocVQA \citep{DocVQA}, we evaluate and train using the prompts from Gemini \citep{Gemini1.5}.

\textbf{TextVQA}
\begin{tcolorbox}
Answer the following question about the image using as few words as possible. Follow these additional instructions:\newline
-Always answer a binary question with Yes or No.\newline
-When asked what time it is, reply with the time seen in the image.\newline
-Do not put any full stops at the end of the answer.\newline
-Do not put quotation marks around the answer.\newline
-An answer with one or two words is favorable.\newline
-Do not apply common sense knowledge. The answer can be found in the image.\newline
Question: {question}
\end{tcolorbox}

\textbf{DocVQA}
\begin{tcolorbox}
Give a short and terse answer to the following question. Do not paraphrase or reformat the text you see in the image. Do not include any full stops. Just give the answer without additional explanation.\newline
Question: \{question\}
\end{tcolorbox}

We use the stop words \texttt{Question}, \texttt{User}, \texttt{<end\_of\_utterance>} and the EOS token to stop a generation.

\subsubsection{Detailed performance on MMMU}

The detailed performance of Idefics3 across each category of MMMU \citep{MMMU} is present in Table \ref{table:detail_performance_mmmu}.

\vspace{0.5cm}

\begin{table}[ht]
\small
\centering
\begin{tabular}{cc}
\toprule
\textbf{MMMU category} & \textbf{Score} \\ \midrule
\textit{Overall} & 46.6 \\
\textit{Accounting} & 33.3 \\
\textit{Agriculture} & 56.7 \\
\textit{Architecture and Engineering} & 33.3 \\
\textit{Art} & 56.7 \\
\textit{Art Theory} & 76.7 \\
\textit{Basic Medical Science} & 50.0 \\
\textit{Biology} & 36.7 \\
\textit{Chemistry} & 40.0 \\
\textit{Clinical Medicine} & 53.3 \\
\textit{Computer Science} & 50.0 \\
\textit{Design} & 73.3 \\
\textit{Diagnostics and Laboratory Medicine} & 43.3 \\
\textit{Economics} & 40.0 \\
\textit{Electronics} & 40.0 \\
\textit{Energy and Power} & 36.7 \\
\textit{Finance} & 40.0 \\
\textit{Geography} & 50.0 \\
\textit{History} & 56.7 \\
\textit{Literature} & 80.0 \\
\textit{Manage} & 47.7 \\
\textit{Marketing} & 53.3 \\
\textit{Materials} & 26.7 \\
\textit{Math} & 26.7 \\
\textit{Mechanical Engineering} & 33.3 \\
\textit{Music} & 26.7 \\
\textit{Pharmacy} & 53.3 \\
\textit{Physics} & 26.7 \\
\textit{Psychology} & 53.3 \\
\textit{Public Health} & 46.7 \\
\textit{Sociology} & 56.7 \\
\textit{Art \& Design} & 58.3 \\
\textit{Business} & 42.7 \\
\textit{Health \& Medicine} & 49.3 \\
\textit{Humanities \& Social Science} & 61.7 \\
\textit{Science} & 36.0 \\
\textit{Tech \& Engineering} & 39.5 \\
\bottomrule
\end{tabular}
\vspace{0.7em}
\caption{Detailed performance of Idefics3 across each category of MMMU. \citep{MMMU}.}
\label{table:detail_performance_mmmu}
\end{table}

\end{document}